%% file: main.tex
\documentclass[conference]{IEEEtran}
\renewcommand{\figurename}{Fig.}

\usepackage[dvips]{graphicx}
\usepackage{amsmath,amsfonts,amssymb}
\usepackage{algorithm}
\usepackage{algorithmic}
\usepackage{url}        
\usepackage{amsmath}    
\usepackage{hyperref}
\hypersetup{draft}
\usepackage{tikz}
\usepackage[labelsep=period,font=footnotesize]{caption}
\usepackage{balance}
\usepackage{parskip}
\begin{document}
\title{Ground Profile Recovery from Aerial 3D LiDAR-based Maps}
\date{}

\author{
\IEEEauthorblockN{Adelya~Sabirova, Maksim~Rassabin, Roman~Fedorenko and Ilya Afanasyev}
\IEEEauthorblockA{\textit{Innopolis University} \\ Innopolis, Russia \\
\{a.sabirova, m.rassabin, r.fedorenko, i.afanasyev\}@innopolis.ru}}

\maketitle


\input{Sections/abstract}
\input{Sections/introduction}
\input{Sections/methodology}
\input{Sections/system}

\input{Sections/mapping}
\input{Sections/processing}

\input{Sections/results}
\input{Sections/conclusion}
\balance
\bibliography{main}
\bibliographystyle{ieeetr}
\end{document}

%% file: Sections/abstract.tex
\begin{abstract}
The paper presents the study and implementation of the ground detection methodology with filtration and removal of forest points from LiDAR-based 3D point cloud using the Cloth Simulation Filtering (CSF) algorithm. The methodology allows  to recover a terrestrial relief and create a landscape map of a forestry region. As the proof-of-concept, we provided the outdoor flight experiment, launching a hexacopter under a mixed forestry region with sharp ground changes nearby Innopolis city (Russia), which demonstrated the encouraging results for both ground detection and methodology robustness.
\end{abstract}

%% file: Sections/introduction.tex
\section{Introduction}
\label{introduction}

The recent success in development of low power and light weight 3D LiDARs for mobile applications have allowed to acquire precise data about terrestrial relief and built high-resolution digital terrain models (DTMs) for ground detection, landscape simulation, forestry monitoring, land-cover classification and many other applications \cite{yan2015, mallet2016, serifoglu2018, pierzchala2018, pfeifer2018, toth2018}. 

Since creation of 3D LiDAR enabled the acquisition of 3D point clouds in forests and a detailed 3D analysis of forest structures. Depending on the density of points, some methods working on the stand and plot level have become operational, providing valuable forest parameters for the inventory. Therefore, there are a lot of investigations developed for forest technologies based on 3D point cloud from terrestrial LiDAR \cite{bauwens2016, michez2016, aijazi2017, itakura2018, ferrara2018, van2018}. 
Some studies are based on deletion tree points from 3D point cloud to recover and detect individual trees, their shapes, canopy, tree height, biomass, leaf inclination angle for which lots of software was developed \cite{ferraz2016, zhao2018, goldbergs2018, chakraborty2019}. Since geological organizations are interested in individual tree detection from point cloud, the paper \cite{itakura2018} presents an automatic individual tree detection through analysis of forest terrestrial relief from point cloud that processed due to terrestrial LiDAR. As results, they achieve tree detection with high accuracy and each tree canopy segmentation. The research \cite{ferrara2018} describes the automatic approach for wood-leaf separation from point cloud that processed also due to terrestrial relief using density based clustering algorithm. As a result, it was proposed the method of wood component extraction from 3D point clouds for broad leaved non-deciduous trees. Unlike most of the published algorithms that detect individual trees from a LiDAR-derived raster surface, the authors \cite{zhang2015} worked directly with the LiDAR point cloud data to separate individual trees and estimate tree metrics.

Some investigations for extracting individual trees and forest segmentation connect airborne and terrestrial measurements \cite{lamprecht2017, hancock2017, schneider2019}. Thus, the authors \cite{lamprecht2017} propose a method which uses a random forest classifier to estimate the matching probability of each terrestrial-reference and aerial detected tree pair within a terrestrial sample plot to aerial detected trees. However, there appeared many recent researches related to detecting individual trees and forest patch delineations from airborne laser scanning (ALS) point clouds based on \cite{kandare2016, kobler2007, puliti2019, bruggisser2019}. Aimed at error reduction and accuracy refinement, the research \cite{chen2018} presents an adaptive mean shift-based clustering scheme aided by a tree trunk detection technique to segment individual trees and estimate tree structural parameters based solely on the airborne LiDAR data.  \cite{dai2018} developed an algorithm to segment individual trees from the small footprint discrete return airborne LiDAR point cloud. Method works by segmenting trees individually in sequence from the point cloud by taking advantage of the relative spacing between trees. \cite{lee2017} develop a 3D tree delineation method which uses graph cut to delineate trees from the full 3D LiDAR point cloud, and also makes use of any optical imagery available. 

Thanks to flying unmanned aerial vehicles (UAVs) at low altitudes, high-density point clouds for accurate representation of a terrain relief are generated \cite{zahawi2015, szabo2018}. Moreover, UAV-based laser scanning is proposed for enabling automated 3D mapping in forests with Simultaneous Localization and Mapping (SLAM) in a combination with the robust graph optimization after loop closures, called graph-SLAM, as a component of forest monitoring \cite{pierzchala2018}.

The ground detection task is very important itself for land classification in terms of cost and suitability for construction and agriculture, since it strongly depends on ground flatness and presence of height changes. In addition, after removing ground points, the forestry region can be separately investigated in terms of tree biomass and forest structure.
To generate DTMs, ground and non-ground measurements have to be separated from the LiDAR point clouds by filtering methods, which remove points of ground objects and extract ground points. This is an important process for most environment modeling applications and is performed using various types of commercial and non-commercial software.
Thus, the author \cite{vosselman2000} proposed a slope method for filtering laser data. A common assumption of slope-based algorithms is that the change in slope usually occurs in the neighborhood gradually, while the change in slope between trees and the ground is large. 
Rashidi and Rastiveis \cite{rashidi2018} also proposed the Slope and Progressive Window Thresholding (SPWT) for ground filtering of LiDAR data, which is based on both multiscale and slope methods.
The limitation of slope-based algorithms is acquiring an optimal slope threshold that can be applied to terrain with different topographic features. The paper \cite{susaki2012} proposes an adaptive filter to overcome such limitation. The paper \cite{zhang2016} presents the Cloth Simulation Filtering (CSF) algorithm as an effective filtering method, which needs a few easy-to-set integer and Boolean parameters to achieve high accuracy of separating point clouds into ground and non-ground measurements (the achieved average total error was about $5\%$). 

In this work, the CSF algorithm was chosen as the base algorithm to extract ground points from LiDAR point cloud according to the paper \cite{serifoglu2018}, where CSF was compared with the ground filtering algorithms of several softwares widely used for processing ALS point clouds and demonstrated the best filtering results. Therefore, we apply the methodology of offline ground detection from LiDAR-based 3D point cloud with the CSF algorithm in \textit{CloudCompare} software, implementing a terrestrial relief extraction with the preliminary outlier removal and point cloud normalization. As the proof-of-concept, we provided the outdoor flight experiment with \textit{DJI M600Pro Hexacopter}, launching the UAV under a mixed forestry region with sharp ground changes nearby Innopolis city (Russia) and acquiring 3D data with the stabilized Velodyne VLP16 LiDAR. The experiment demonstrated that applying the proposed methodology allows achieving encouraging results for ground profile detection, having about $20\%$ of ground points from the total amount of the 3D point cloud from LiDAR rawdata for the flight under the mixed forest in the middle of autumn. Although, the total changes in relief height was about 40 meters, the proposed methodology of ground filtering shown the robustness to these conditions.

The technological contribution of the paper consists in the integration of airborne 3D LiDAR data processing solutions into the applied methodology for ground detection using filtration and removal of forest points from three-dimensional point cloud based on the Cloth Simulation Filtering (CSF) algorithm. The main limitation of the research is that although the reconstructed ground profile obtained during flight experiments is credible, it must be metrologically verified using other more accurate and proven technologies.

The rest paper is structured as follows. Section \ref{methodology} introduces the methodology for LiDAR mapping from a copter with ground detection. Section \ref{system} describes our hexacopter configuration and software used. Sections \ref{mapping} and \ref{processing} outline aerial mapping system and the logics of ROS-based sensor data processing correspondingly. Section \ref{results} presents the experimental results with comparative analysis for point cloud data before and after filtering. Finally, we conclude in Section \ref{conclusion}. 

%% file: Sections/methodology.tex
\section{Methodology}
\label{methodology}

\begin{figure}[!htbp]
    \centering
    \includegraphics[width=0.3\linewidth]{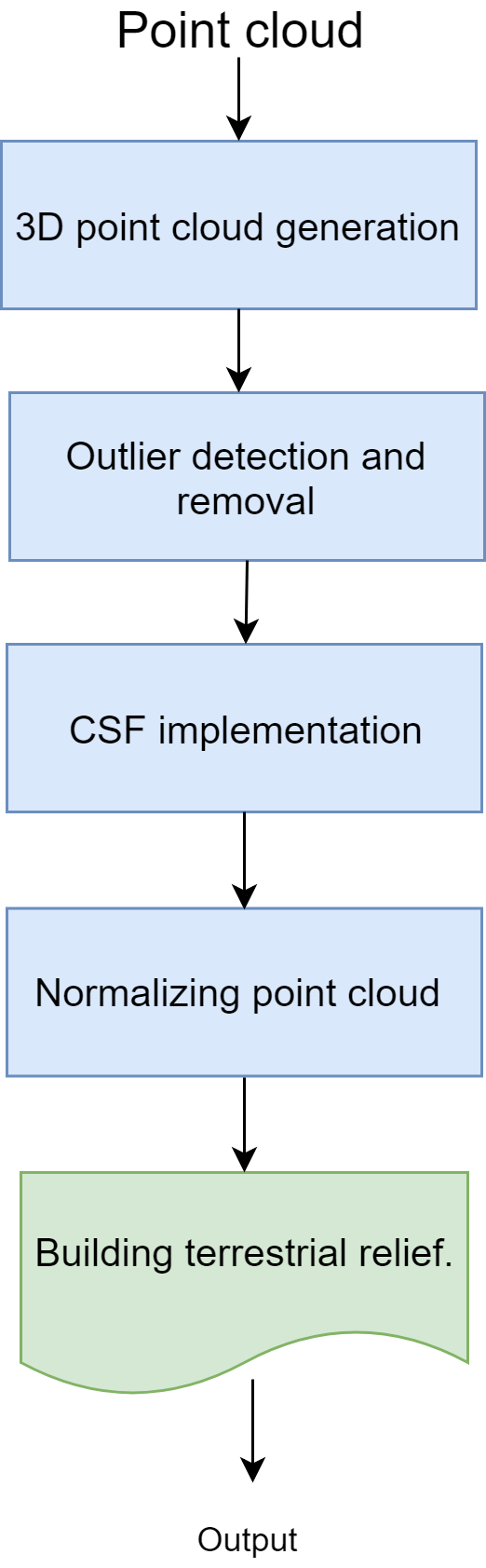}
    \caption{The block-scheme of the proposed methodology for ground detection using filtration and removal of forest points from airborne 3D LiDAR-related point cloud based on the Cloth Simulation Filtering (CSF) algorithm}
    \label{fig01}
\end{figure}

The methodology of ground detection from LiDAR-based 3D point cloud consists of 6 steps, the schematic overview of which is presented in Fig. \ref{fig01}. During the stage of LiDAR data acquisition the forestry region was investigated with Velodyne LiDAR mounted on a hexacopter with recording LiDAR dataset to the UAV memory for the following offline processing. 
The chosen forestry region is located nearby Innopolis city, in the Republic of Tatarstan, Russia. In Section \ref{mapping}, all details regarding to the outdoor experiment are presented. 
	
\subsection{The first step: 3D point cloud generation} 
3D point cloud was created offline from recorded LiDAR data with ROS packages. The Section \ref{processing} describes sensor data processing and point cloud generation.
	
\subsection{The second step: Outlier detection and removal} 
As known, the 3 sigma-rule is a widely used heuristic for outlier detection of some statistical hypotheses whose test statistics are normalized that can be applied to geodetic data adjustment as shown in the paper \cite{lehmann2013}. In our case, we could programmatically choose the sigma coefficient from the array [1..3]. In the results of practical calculations we get the best result with sigma coefficient that is equal to 1.2. Moreover, since we have the huge point cloud dataset in small geographical region, we selected 20 nearest points for calculation of mean distances.

\subsection{The third step: Cloth Simulation Filtering (CSF) implementation}
\begin{figure}[!htbp]
    \centerline{\includegraphics[width=\linewidth]{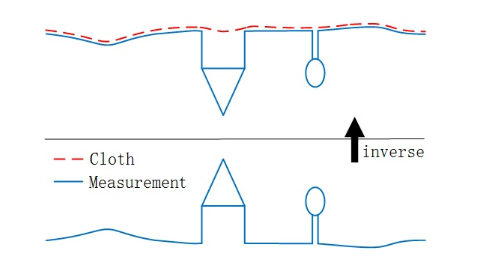}}
    \caption{The illustration of the Cloth Simulation Filtering (CSF) algorithm. The original point cloud is turned upside down, and then a simulated fabric falls on the inverted surface from above, dividing the point clouds into ground and non-ground parts. \copyright \! Courtesy of Zhang, et al. \cite{zhang2016}}
    \label{fig02}
\end{figure}

At using the Cloth Simulation Filtering (CSF) method, 3D point cloud is splitted into two parts: ground and environment points. Taking a closer look at CSF method for filtering ground points \cite{zhang2016}, we recognize that the ground detection and relief recover can be processed by implementation filtering of point cloud based on Cloth Simulation. This algorithm is based on separating point clouds into ground and non-ground measurements, using a simulation. For this reason, the original point cloud is turned upside down, and then a cloth falls on the inverted surface from above. Analyzing the interactions between the nodes of the cloth and the corresponding LiDAR points, the final shape of the cloth can be determined and used as a base to classify the original points into ground and non-ground parts \cite{szabo2018}. The main idea of the CSF algorithm is illustrated in Fig. \ref{fig02}. 
	
The CSF algorithm applies four user-defined parameters: a grid resolution (GR), which represents the horizontal distance between two neighboring particles; a time step (dT) that controls the displacement of particles from gravity during each iteration; a rigidness (RI), which describes the cloth rigidness; and a steep slope fit factor (ST), which is an optional parameter for indicating whether the post-processing of handling steep slopes is required or not. 
    
According to \cite{zhang2016} the optimal value of the time step (dT) is equal to 0.65 that is achieved by the number of maximum iterations of 500. The grid resolution number (GR) is set to 0.1. Although, to increase the resolution of the filtered point cloud it is necessary to choose the minimal grid resolution value. However, since the configuration of the system does not allow to set value less than 0.1, the results presented in the section \ref{results} are based on this value. The parameter of the steep slope fit factor is Boolean and consists of two values "true" or "false" that denote the existence or absence of a steep slope. Since we have a steep slope in the point cloud, this factor is set to 'true' value that provides the additional post-processing. The rigidness is set to 2 that means the presence of terraced slopes in the relief because of the character of the investigated forestry area with with ravines. The rigidness parameters of 1 and 3 define areas with high steep slopes and gentle slopes (flat surface) respectively.
	
\subsection{The forth step: Point cloud normalization and Building terrestrial relief} 
To normalize point cloud which represents terrestrial relief for building surface. Finally, to build surface that represents terrestrial relief. 

For normalization of point cloud and surface plotting we should calculate normal vectors as a base for surface. The surface, which is normal to a point, estimates the surrounding neighborhood points that support the point (also called k-neighborhood). It is known that the best choice of k depends on the data and, as a rule, larger values of k reduces noise influence on the classification, but make boundaries between classes less clear \cite{everitt2011}. Then after choosing groups of neighbors it is necessary to build a plane, plotting normals to each of plane. For this part of our task it is sufficient to choose default parameters for normals plotting:

\begin{itemize}
        \item k-neighborhood = 100;
        \item number of planes = 1000;
        \item Accumulator steps = 15;
        \item Number of rotations = 5;
        \item Tolerance angle = 90 deg.;
        \item Neighborhood size for density estimation = 5.
\end{itemize}

%% file: Sections/system.tex
\section{System Setup}
\label{system}

For filtration and deletion of the trees' points with the consecutive ground points' detection from 3D point cloud of LiDAR-data we used the following system configuration, containing Hexacopter, \textit{Velodyne} LiDAR and a Laptop with installed \textit{Ubuntu 16.04 (Xenial Xerus)}. 

In our outdoor experiments we used \textit{DJI M600Pro} Hexacopter with the stabilized Velodyne VLP16 LiDAR. Although this LiDAR was mainly developed for transport vehicles as a low-cost collision avoidance sensor, nevertheless it has been widely used for unmanned aerial vehicles (UAVs) \cite{jozkow2016, hening2017, guo2017, pierzchala2018}, indoor mapping platforms \cite{yagfarov2018}, mobile robots \cite{meng2016} and autonomous vehicle designs \cite{nobili2015} as the primary mapping device for acquiring high resolution 3D models. The accuracy and stability analysis of the VLP16 laser scanner was studied at the work \cite{glennie2016}.
The hardware and software system configuration is presented in the Table \ref{tab1}.

\begin{table}[!htbp]
\caption{\label{tab1}\fontsize{10}{12}\fontfamily{ptm}\selectfont \textsc{Computational System Specification}}
\centering
\begin{tabular}{|c|c|}
		\hline
 		\textbf{System} & \textbf{Configuration} \\
 		\hline
		Laptop & $LENOVO^{TM} \quad ideapad \quad 320S$ \\
		\hline
		Operating System & x64, Windows 10 \\
		\hline
		RAM & 8 GB \\
		\hline
		CPU & $Intel^R$ $Core^{TM} \quad i5$ \\
		\hline
\end{tabular}

\end{table}

To process points from \textit{Velodyne} LiDAR and build the whole point cloud by a filtration algorithm, we utilized Robot Operating System (ROS) framework in Kinetic Kame version. 
Additionally, the 3D point cloud processing software of \textit{CloudCompare 2.9.1 Omnia}  is an open source software for 3D point cloud processing that
was used for filtering ground and non-ground points with CSF filtering, recovering terrestrial relief. The main advantage of this algorithm is that it uses very few easy-to-set parameters.
Alternatively, the Cloth Simulation Filtering (CSF) can be utilized in MATLAB with the CSF library available on the \textit{MathWorks} website.

\begin{figure}[!htbp]
    \centerline{\includegraphics[width=\linewidth]{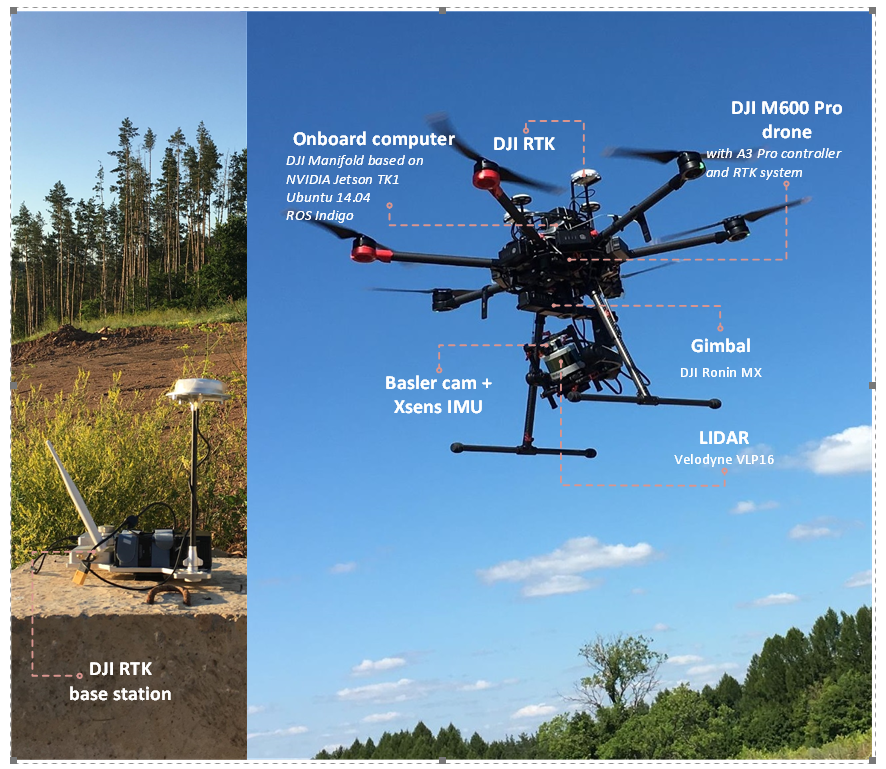}}
    \caption{The experimental aerial mapping system based on the \textit{DJI M600Pro} Hexacopter with the  stabilized \textit{Velodyne VLP16} LiDAR}
    \label{fig04}
\end{figure}

%% file: Sections/mapping.tex
\section{Aerial mapping system}
\label{mapping}

Aerial mapping system was implemented in research \cite{fedorenko2018}, which contains example of usage of point cloud map, gathered from drone, for UGV traversable path planning. We utilize the same approach for data gathering, but use it for another application, namely trees filtering and ground surface estimation. We used \textit{DJI Matrice 600 Pro} developer drone with \textit{A3 Pro} flight controller equipped with real-time kinematic (RTK) system used to enhance the precision of position data derived from satellite-based positioning systems. Sensor module is mounted on \textit{DJI Ronin MX} gyrostabilized gimbal and includes 3d LiDAR and camera with optional \textit{Xsens} IMU.

Onboard computer with \textit{ROS Indigo} is used to process navigation and sensor data. We use \textit{DJI Onboard SDK} to communicate with A3 flight controller via UART interface. LiDAR interface is Ethernet, camera and optional IMU are connected via USB. At current experiment we saved all data to \textit{ROS bag} file during flight for further processing on laptop.

We used \textit{DJI GroundControl iOS} app for mission planning. Thus, the flight was executed at the height of about 50m in a preliminary set flight zone within the boundaries assigned by the hexacopter GPS coordinates. During the flight operator could see RViz GUI window with camera and LiDAR data visualization, which is done onboard and transmitted from computer HDMI port by \textit{DJI Lightbridge} transmission system.

\begin{figure}[!htbp]
    \centerline{\includegraphics[width=\linewidth]{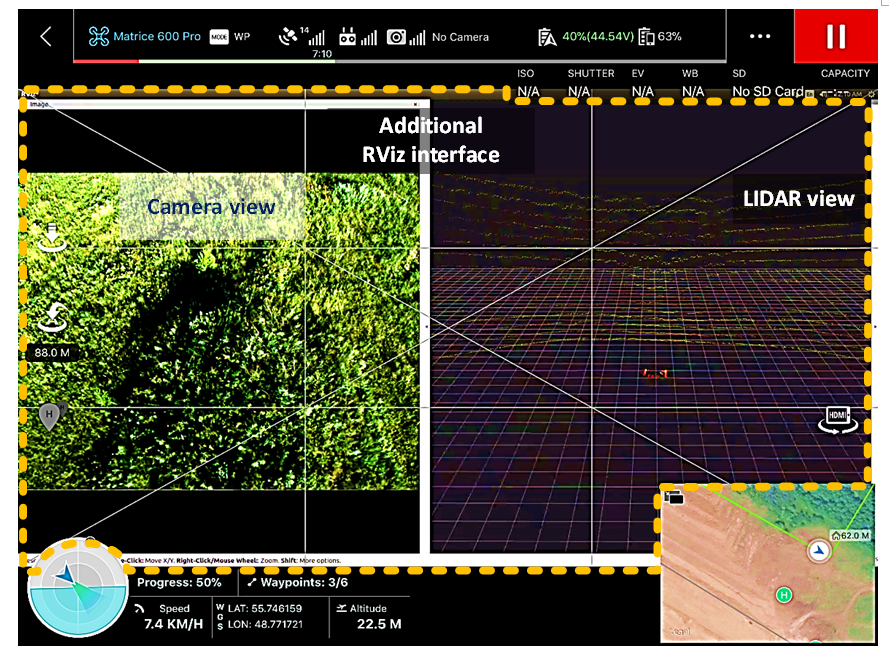}}
    \caption{\textit{DJI GroundControl} app with \textit{RViz GUI}}
    \label{fig06}
\end{figure}

%% file: Sections/processing.tex
\section{Sensor data processing}
\label{processing}

\begin{figure}[!htbp]
\centerline{\includegraphics[width=\linewidth]{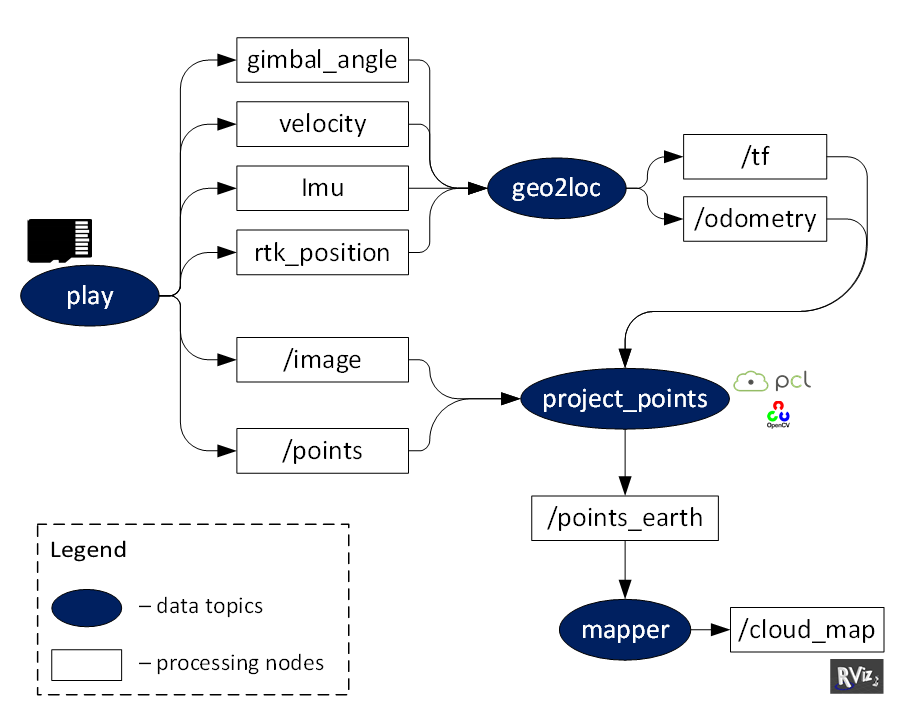}}
\caption{\textit{Rosbag} processing dataflow for 3d  point  cloud  map construction  by  processing  a  sequence  of  sensor  data  from the onboard Hexacopter dataset}
\label{fig07}
\end{figure}

After completion of mapping mission, we process \textit{rosbag} file to get point cloud map of environment. The Fig. \ref{fig07} shows the dataflow of the environment and 3d point cloud map constructing by processing a sequence of sensor data from the dataset. The dataset consists of navigation data, namely RTK position, IMU data, velocity data, gimbal angles data, and sensors data, namely LiDAR packet (converted to point cloud) and compressed images from camera. The \textit{Geo2loc} node forms \textit{/tf} coordinate frames tree and local ENU frame \textit{/odometry} data for further projection of sensor data from current drone body frame to earth ENU frame, which is done by \textit{project\_points} node. The \textit{Project\_points} node could optionally utilize camera data to colorize LiDAR point cloud in camera field of view. The \textit{Mapper} node combines all point clouds from separate LiDAR measurements in united cloud map, which is saved from ROS topic to \textit{.pcd} and \textit{.ply} files for further processing.

%% file: Sections/results.tex
\section{Experimental results}
\label{results}
\begin{figure}[!htbp]
    \centerline{\includegraphics[width=\linewidth]{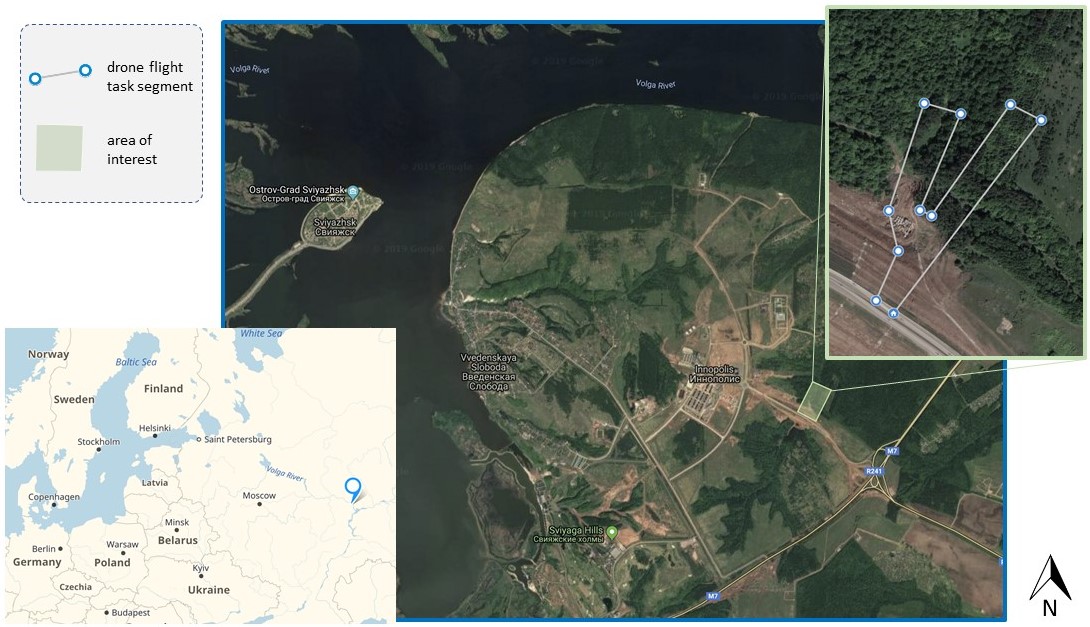}}
    \caption{The experimental mapping area location nearby Innopolis  city, the Republic of Tatarstan, Russia}
    \label{fig_area_map}
\end{figure}

\begin{figure}[!htbp]
    \centerline{\includegraphics[width=\linewidth]{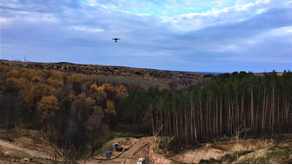}}
    \caption{The photo of the mapping area (a mixed forest nearby Innopolis city, the Republic of Tatarstan, Russia) investigated by flight experiments of \textit{DJI M600Pro} Hexacopter with the stabilized \textit{Velodyne VLP16} LiDAR}
    \label{fig_area_photo}
\end{figure}

\begin{figure}[!htbp]
\centerline{\includegraphics[width=0.9\linewidth]{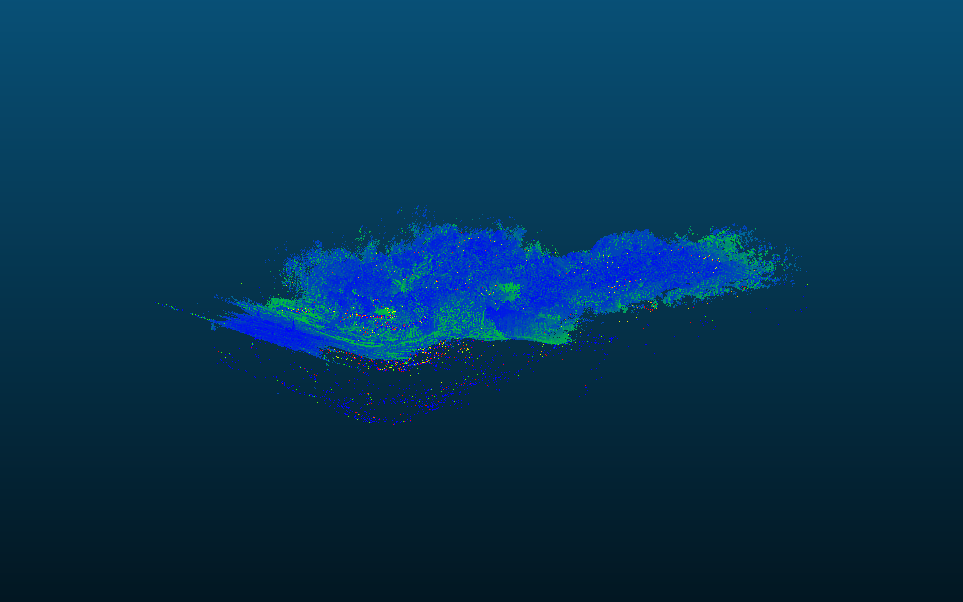}}
\caption{The initial point cloud of the forestry region from LiDAR rawdata (after ROS processing, but before filtration)}
\label{fig08}
\end{figure}

The mixed forest area located nearby Innopolis city, the Republic of Tatarstan, Russia, as shown in the Fig. \ref{fig_area_map} was scanned during the UAV flight experiment in the middle of October 2018. The photo of data gathering flight at the start of the experiment is shown in the Fig.\ref{fig_area_photo}.
The initial point cloud of the forestry area obtained from LiDAR rawdata (after ROS processing, but before filtration) is presented in the Fig. \ref{fig08}.

\begin{figure}[!htbp]
\centerline{\includegraphics[width=0.9\linewidth]{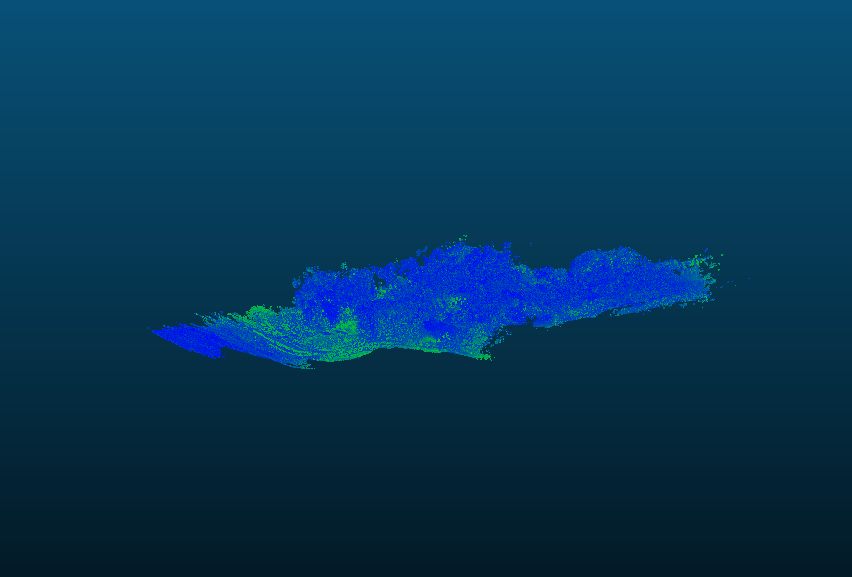}}
\caption{3D point cloud of the forestry region without outliers}
\label{fig09}
\end{figure}

After receiving point cloud, several experiments was produced to find optimal values. As described in the Section \ref{methodology} the optimal parameter of the standard deviation was set to 1.2, and the number of nearest points was selected to 20. In the Fig. \ref{fig09} point cloud is presented without noise. 

\begin{figure}[!htbp]
\centerline{\includegraphics[width=0.9\linewidth]{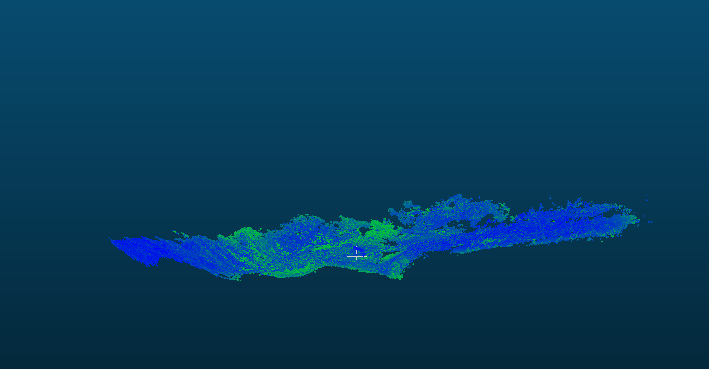}}
\caption{Point cloud after filtering with the Cloth Simulation Filtering (CSF) algorithm that reconstructs the ground points}
\label{fig10}
\end{figure}

The next step is 3D point cloud filtering with Cloth Simulation algorithm considered in the Section \ref{methodology}-C. According to the algorithm, the threshold of 0.6 splits the points into two groups: ground and non-ground. The Fig. \ref{fig10} shows the results after Cloth Simulation Filtering. If we zoom in the figure, we can see that the point clouds have no dense structure in some parts. This issue is connected to ground points which are not reflected back to the LiDAR. Nevertheless, the described procedure allows to obtain a satisfactory contour. The Table \ref{tab2} presents the quantitative analysis for points in each point cloud groups that were generated, and amount of points in both outliers and LiDAR rawdata. The table demonstrates that applying the proposed methodology allows achieving the ground profile detection, having about $20\%$ of ground points from the total amount of the 3D point cloud from LiDAR rawdata for the flight under the mixed forest in the middle of autumn. It illustrates the robustness of the proposed methodology of the ground detection.

\begin{figure}[!htbp]
    \centerline{\includegraphics[width=0.9\linewidth]{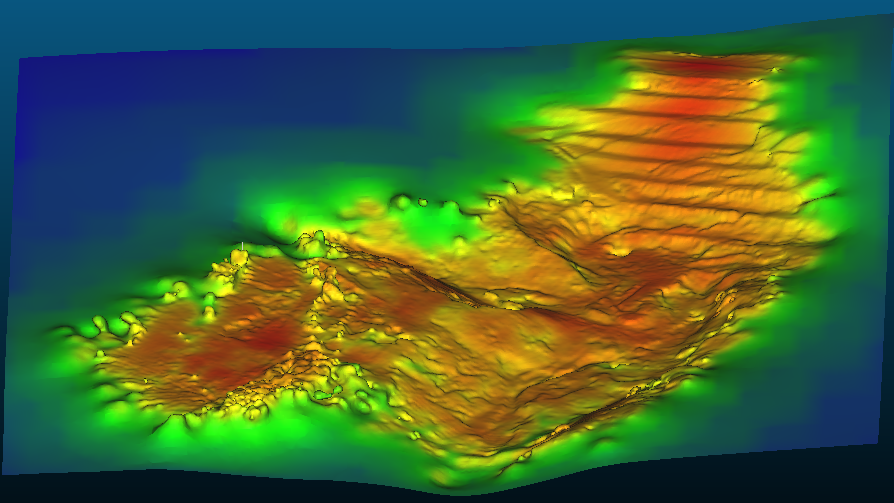}}
    \caption{The profile of the forestry region recovered from point cloud by the algorithm of plotting surface based on normals}
    \label{fig11}
\end{figure}

\begin{figure}[!htbp]
    \centerline{\includegraphics[width=0.9\linewidth]{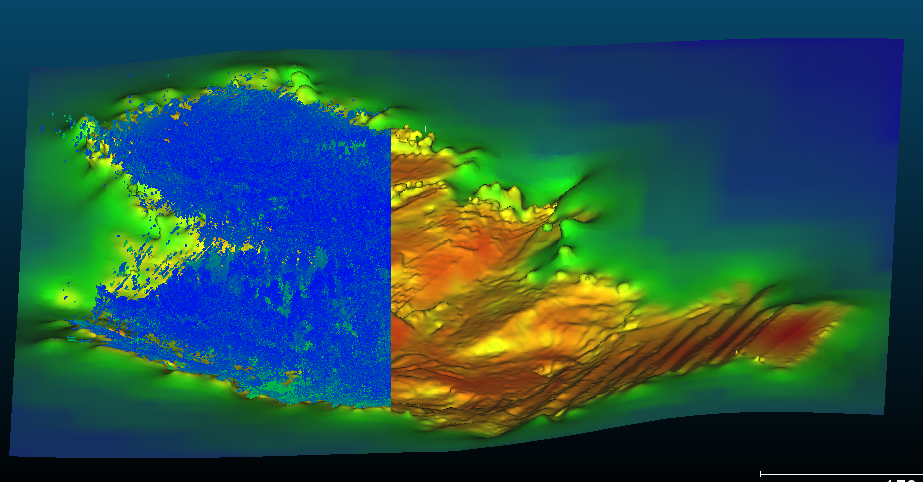}}
    \caption{Original point cloud superimposed on a half of the surface (representing a comparison between ground  models  before  and  after 3D LiDAR data filtering)}
    \label{fig12}
\end{figure}

After applying default algorithm for plotting surface based on normals we get surface. The Fig. \ref{fig11} represents such surface. The method added some points around ground points 
to create a complete square outline. The main interest is the points 
in the middle of the recovered terrain, which show that the surface is based on the maximum number of points remaining after filtering.
The comparison of ground models before and after filtering is shown in Fig. \ref{fig12}.

\begin{figure}[!htbp]
    \centerline{\includegraphics[width=0.9\linewidth]{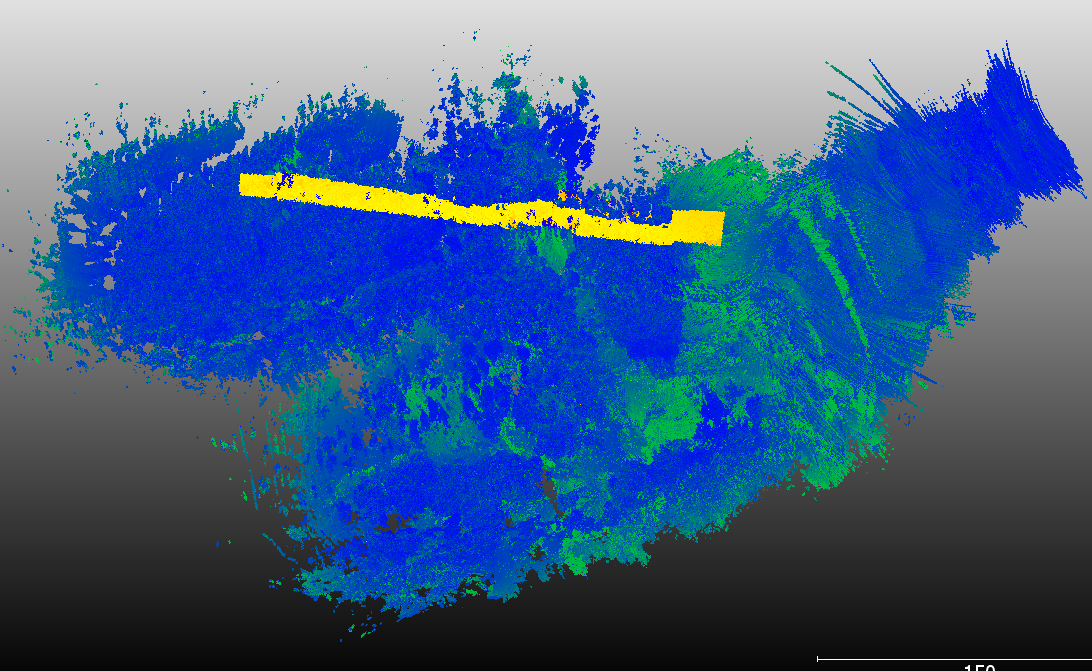}}
    \caption{The region of point cloud (dark points) with the cut (the light line) for exploring the height profile using a zoom}
    \label{fig13}
\end{figure}

\begin{figure}[!htbp]
    \centerline{\includegraphics[width=0.9\linewidth]{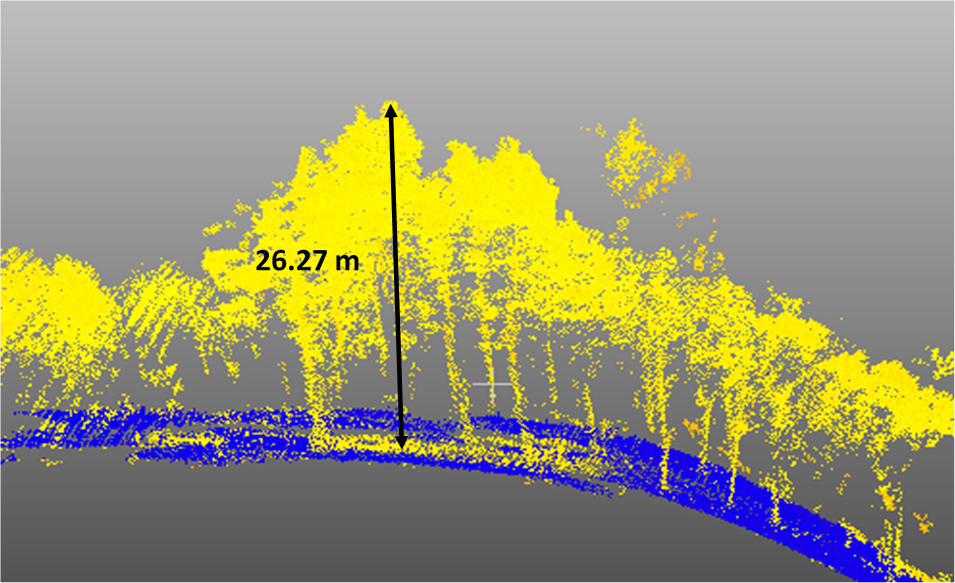}}
    \caption{The relief profile changes in the cut region indicated in the Fig. \ref{fig13}, with the marked height of the tree}
    \label{fig14}
\end{figure}

\begin{figure}[!htbp]
    \centerline{\includegraphics[width=0.9\linewidth]{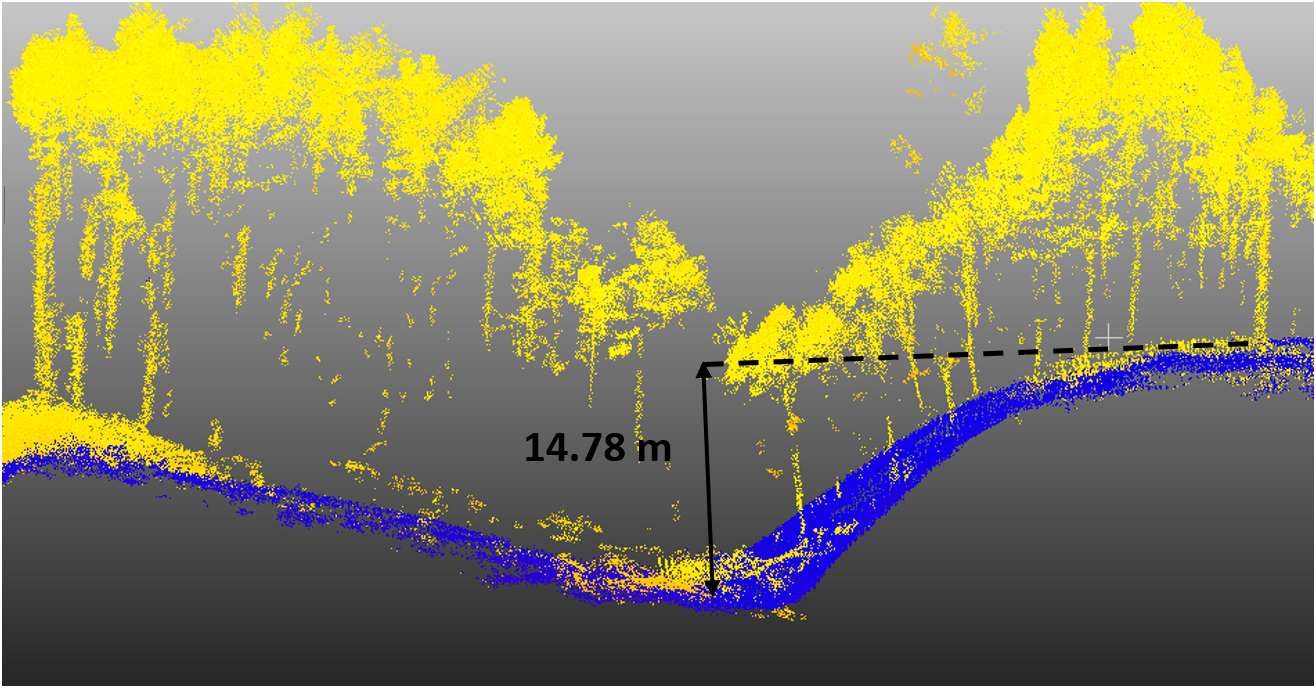}}
    \caption{The profile of the ravine relief with trees in the cut region, indicated in Fig. 13, with the marked delta of the ground height in the cut}
    \label{fig15}
\end{figure}

Also we can say that the proposed methodology is applicable to rather difficult conditions, because, for example, in the cut region under consideration in the Fig. \ref{fig13} there are pines with a dense coating of needles and high crowns. What is more, the Fig. \ref{fig14} and Fig. \ref{fig15} demonstrate the large differences in relief profile changes, which undoubtedly complicates the algorithm implementation for ground filtering. Although, the total changes in relief height is about 40 meters, nevertheless, the proposed methodology shown robustness to these conditions.

\begin{table}[!htbp]
\caption{\label{tab2}\fontsize{10}{12}\fontfamily{ptm}\selectfont \textsc{The quantitative analysis for points in each point cloud groups}}
\centering
\begin{tabular}{|c|c|c|}
		\hline
 		\textbf{Point Cloud types} & \textbf{Amount of points} & \textbf{Percent} \\
		\hline
		Point Cloud without outliers & 20 805 983 & 93.75 \\
 		\hline
		Outliers & 1 387 160 & 6.25 \\
		\hline
		Ground Point Cloud & 4 480 098 & 20.19 \\
		\hline
		Trees Point Cloud & 16 325 885 & 73.56 \\
		\hline
		\textbf{Point Cloud from Rawdata} & \textbf{22 193 143} & \textbf{100} \\
		\hline
\end{tabular}
\end{table}

%% file: Sections/conclusion.tex
\section{Conclusions}
\label{conclusion}

 In this research, we technologically contributed into the development and implementation of methodology of offline ground detection from airborne 3D LiDAR-related dataset by integration of data processing solutions using filtration and removal of forest points from three-dimensional point cloud based on the Cloth Simulation Filtering (CSF) algorithm. 
 
 Such a methodology is quite important and applicable for land classification in terms of cost and suitability for construction and agriculture (especially in suburbs nearby cities and villages), since it strongly depends on ground flatness and presence of ravines and hills.  

We realized this methodology in \textit{CloudCompare} software, executing a terrestrial relief extraction with the preliminary outlier removal and point cloud normalization. The applied methodology allowed to recover the terrestrial relief. In addition, after removing ground points, the forestry region can be investigated separately in terms of tree biomass and forest structure. After analyzing the received results of 3D point cloud, the optimal values was found for noise decrease.

As the proof-of-concept, we provided the outdoor flight experiment with \textit{DJI M600Pro Hexacopter}, launching the UAV under a mixed forestry region with sharp ground changes nearby Innopolis city (the Republic of Tatarstan, Russia) and acquiring 3D data with the stabilized Velodyne VLP16 LiDAR. \textit{DJI GroundControl iOS} app was used for UAV mission planning, executing the flight at the height of about 50m under a preliminary set forest zone within the boundaries assigned by the hexacopter GPS coordinates. An onboard computer with Robot Operating System (ROS) was applied to process the UAV autonomous navigation and sensor data acquisition. 
The experiment demonstrated that exploiting the implemented methodology allows achieving encouraging results for ground profile detection, having about $20\%$ of ground points from the total amount of the 3D point cloud from LiDAR rawdata for the flight under the mixed forest in the middle of autumn. Although, the total changes in relief height was about 40 meters, the proposed methodology of ground filtering shown the robustness to these conditions.

In future work, for better estimation of experimental results it is proposed to use the aerial laser scanning for a preselected and geodesically measured forest area, and then, after the 3D point cloud filtering, analyze the accuracy of the algorithm. 